\documentclass[]{Liebert_Author}

\usepackage{amsmath,amssymb}
\usepackage{graphicx}
\usepackage{xcolor}
\usepackage{longtable}
\usepackage{booktabs}
\usepackage{multicol}
\usepackage{multirow}
\usepackage{xspace}
\usepackage{caption}
\usepackage{subcaption}
\usepackage{float}
\usepackage{hyperref}  
\usepackage{comment}
\newtheorem{definition}{Definition}

\newcommand{\ie}{\emph{i.e.}}
\newcommand{\eg}{\emph{e.g.}}
\newcommand{\ea}{\emph{et al.}}

\title{Characterizing SARS-CoV-2 Spike Sequences Based on Geographical
  Location}

\author{Sarwan Ali, Babatunde Bello, Zahra Tayebi and Murray
  Patterson\\
  {Department of Computer Science, Georgia State University}\\
  {Atlanta, GA 30303, USA}\\
}

\date{}

\begin{document}
%----------------------------------------------------------------------

\maketitle


\begin{abstract}
  With the rapid spread of COVID-19 worldwide, viral genomic data is
  available in the order of millions of sequences on public databases
  such as GISAID.  This \emph{Big Data} creates a unique opportunity
  for analysis towards the research of effective vaccine development
  for current pandemics, and avoiding or mitigating future pandemics.
  One piece of information that comes with every such viral sequence
  is the geographical location where it was collected --- the patterns
  found between viral variants and geographical location surely being
  an important part of this analysis.  One major challenge that
  researchers face is processing such huge, highly dimensional data to
  obtain useful insights as quickly as possible.  Most of the existing
  methods face scalability issues when dealing with the magnitude of
  such data.  In this paper, we propose an approach that first
  computes a numerical representation of the spike protein sequence of
  SARS-CoV-2 using $k$-mers (substrings) and then uses several machine
  learning models to classify the sequences based on geographical
  location.  We show that our proposed model significantly outperforms
  the baselines.  We also show the importance of different amino acids
  in the spike sequences by computing the information gain
  corresponding to the true class labels.
\end{abstract}

\section{Introduction}
%----------------------------------------------------------------------

\begin{figure}[!ht]
  \centering
  \includegraphics[scale=0.4,page=1]{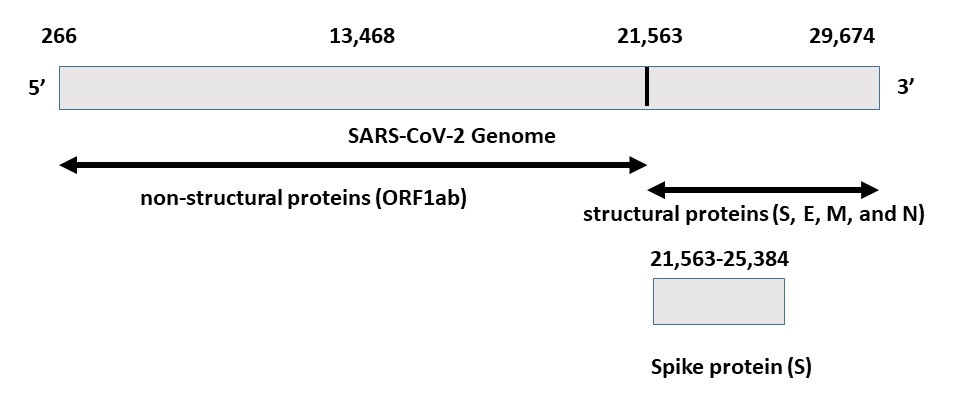}
  \caption{The SARS-CoV-2 genome codes for several proteins, including
    the surface, or spike protein.  The spike protein is composed of
    3821 ($25384-21563$) nucleotides (and one ``stop'' character `*').
    Therefore, the final length of the spike protein is $3822 / 3 =
    1274$ (we divide by $3$ because each amino acid corresponds to $3$
    DNA characters, or codons)~\cite{huang2020structural}.}
  \label{fig_spike_seq_example}
\end{figure}

The adaptability of viruses like SARS-CoV-2, when coupled with a
variety of selection pressures from the various ecosystems, host
immunities and approaches to pharmaceutical intervention provide an
evolutionary environment that leads to the emergence of strains and
variants in different geographical locations.  While SARS-CoV-2 has
spread rather quickly to many parts of the globe since the initial
outbreak in Wuhan at the end of 2019, which led to the COVID-19
pandemic~\cite{wu-2020-new}, it continues to raise global concerns as
the virus persistently evolves and accumulates new mutations.
Consequently, new variants of SARS-CoV-2 have emerged in different
parts of the world: the Alpha variant (B.1.1.17) emerged in the UK,
Beta (B.1.351) in South Africa, Gamma in Brazil, Epsilon in
California, Iota (B.1.526) in New York, Delta (B.1.167.2) and Kappa
(B.1.167.1) in India, to name a few.  All of these variants possess
some mutations that confer increased transmissibility or higher
binding affinity of their spike protein (see
Figure~\ref{fig_spike_seq_example}) to human host ACE2
receptors~\cite{farinholt2021transmission,huang2020structural}.

It is concerning that the longer SARS-CoV-2 has to propagate, its
exposure to wider ranges of immune response attacks across diverse
communities and geographically diverse environments may be incubating
the virus to evolve new variants and strains that are dangerous and
extremely immunologically evasive both locally and globally, as the
pandemic prolongs.  From the point of view of evolution, this is like
giving the virus robust evolutionary room and time to learn, to evolve
adaptations, gain of function, and escapes from host immune arsenal
and attacks.  Sadly, this is gradually the case already, as the
original Wuhan strain is now almost completely replaced by new
variants with different characteristic behaviors and are hence less
responsive to the currently available
vaccines~\cite{korber2020tracking,hu2021characteristics}.  This is why
it is important to characterize different strains and variants of
SARS-CoV-2 based on geographical location, to understand the patterns
of spread in hopes to contain, or at least cope with this virus.

All viruses mutate with time --- RNA viruses particularly do so at a
faster rate. The SARS-CoV-2 is an RNA virus, it however exhibits a
moderately lower rate compared to other RNA viruses like HIV and
influenza due to the possession of a genetic proof-reading mechanism
for correcting errors.  The SARS-CoV-2 genome typically accrues 1 or 2
point mutations (SNVs) in a month.  According to a review, some 12,706
such mutations have so far been detected by researchers since the
advent of the COVID-19 pandemic.  While some changes have neutral
effects, a few that occur in major proteins --- be it, addition,
substitution or deletion --- are critical to viral evolution, genomic
stability, transmissibility, antigenicity, virulence, adaptation and
escape from host immune
response~\cite{pachetti2020emerging,lorenzo2020unique}.  The
SARS-CoV-2 Spike (S) Protein is a key player in the virus life cycle.
The protein is composed of $1274$ amino acids encoded by the S gene of
the virus (see Figure~\ref{fig_spike_seq_example}).  It is the major
target of the neutralizing antibodies from host immune response and
currently available vaccines for COVID-19.  The virus uses the spike
protein to bind the host ACE2 receptor on the cell surface (found
abundantly in airways, lungs, mucous lines and the intestine) which
facilitates the uptake of the virus into host
cells~\cite{lamers2020sars,v2021coronavirus}.  Thus, mutations in the
S gene have reportedly imparted viral pathogenesis, binding activity
of the spike protein to the host, as well as causing conformational
changes in the protein molecule.  For instance, mutation D614G ---
that is, a substitution of Glycine (G) for Aspartate (D) at position
614 --- was found to enhance the viral infectivity and stability of
the SARS-CoV-2 genome, which has been attributed to spike protein
assembly on the virion surface~\cite{korber2020tracking}.

Currently, quite a number of novel variants are being identified by
the US Centers for Disease Control and Prevention (CDC) and the World
Health Organization (WHO)~\cite{CDS_variantDef}.  The variants are
divided into categories such as Variants of Concern (VOCs), Variants
Being Monitored (VBMs), Variants of interest (VOIs), and Variants of
High Consequence (VOHCs). At the time of this study, the VOC was the
Delta variant (B.1.617.2 and AY.1 sublineages)~\cite{CDS_variantDef}.
Since all of these variants are characterized by different spike
protein content~\cite{farinholt2021transmission,huang2020structural},
classification can help us to discover also patterns in the geographic
distribution of these variants.  At the time of this study, the VBMs
comprised Alpha (B.1.1.7, Q.1-Q.8 pango lineage), Beta (B.1.351,
B.1.351.2, B.1.351.3 pango lineage), Gamma (P.1, P.1.1, P.1.2 pango
lineage), Epsilon (B.1.427 B.1.429 pango lineage), Eta (B.1.525 pango
lineage), Iota (B.1.526 pango lineage), Kappa (B.1.617.1 pango
lineage), Zeta (P.2 pango lineage), and MU (B.1.621, B.1.621.1 pango
lineage)~\cite{CDS_variantDef}. There are no VOIs or VOHCs at the
time.

The SARS-CoV-2 still circulates among human populations in different
locations, weather conditions and epidemiological descriptions.  It is
important to investigate how this regional diversity contributes to
viral evolution and emergence of new variants in these regions.
Research suggests possible selective mutations in the SARS-CoV-2
genome --- specific sites which appear more subject to selective
mutation.  Some mutational sites in ORF1ab, ORF3a, ORF8 and N regions
of SARS-CoV-2 reportedly exhibit different rates of
mutation~\cite{wang2020establishment}.  A study involving the analysis
and characterization of samples from COVID-19 patients in different
parts of the world identify 8 novel recurrent mutational sites in the
SARS-CoV-2 genome.  Interestingly, the studies also note changes at
sites 2891, 3036, 14408, 23403, and 28881 to be common in Europe,
while 17746, 17857, and 18060 are common in North
America~\cite{pachetti2020emerging}.  A recent study also identified
the ongoing evolution of SARS-CoV-2 to involve purifying selection,
and that a small number of sites appear to be positively selected.
The work also identifies the spike protein receptor binding domain
(RBD) and a region of nucleocapsid protein to be also positively
selected for substitutions.  The work also highlighted trend in virus
diversity with geographic region and adaptive diversification that may
potentially make variant-specific vaccination an
issue~\cite{rochman2021ongoing}.

Given all of the novel SARS-CoV-2 variants and strains that have
emerged from different geographical regions of the world, we need to
investigate this connection to the spread of the virus, \eg, weather
factors possibly play a systematic
role~\cite{segovia-2021-atmospheric,pezzotti2016approximated}.  There
is also diversity of immune system across the human population.
Genomic variations only cause 20--40\% of this immune system
variation, while the rest 60--80\% is accounted for by age,
environmental factors, such as where we live and our neighbors,
cohabitation and chronic viral infections, etc.  Immune response is
also known to show intra-species variation~\cite{liston2016shaping}.
There is an ongoing evolutionary arms-race between host and pathogens
they are exposed to which constantly changes the host anti-pathogen
attack and in turn causes the pathogen to refine or adjust its escape
from host immune attack~\cite{liston2016shaping,brodin2015variation}.
This is constantly taking place, with the virus under evolutionary
pressure and natural selection to propagate the virus with the highest
fitness.  It may be complex to characterize how each factor
contributes to this variation.  The immune system variation is
possibly an important driver on how new variants of SARS-CoV-2 are
regionally emerging with positive selections for escaping immune
neutralization, increased infectivity and transmissibility, as
observed recently.

Classification of the SARS-CoV-2 Spike protein sequences based on
geographical location of emergence is therefore an important and
informative exploration for possible unique patterns, trends and
distribution.  The SARS-CoV-2 spike protein must interact chemically
with the host receptor molecule, ACE2 for cellular uptake.  Since
millions of spike sequences are available now on public databases such
as GISAID, classifying those sequences becomes a \emph{Big Data}
problem.  When dealing with big data, scalability and robustness are
two important challenges.  Some algorithms are robust while other
scale well, but give poor predictive performance on larger datasets.
The author of~\cite{ali2021spike2vec} proposed a scalable approach,
called Spike2Vec, which is scalable to larger sized datasets.  When
there is some structure (natural clustering) in the data, Spike2Vec is
proven to be useful as compared to one-hot
embedding~\cite{ali2021spike2vec}.  However, we show in this paper
that Spike2Vec does not always work in all types of scenarios.  To
further improve the results of Spike2Vec and that of one-hot
embedding, we use a neural network (NN) model.

In this paper, we propose to use a simple sequential convolutional
neural network along with a $k$-mers based feature vector
representation for classifying the geographical locations of COVID-19
patients using spike protein sequences only.  Our contributions in
this paper are the following:
\begin{enumerate}
\item We show that the neural network model is scalable on a high
  volume of data and significantly outperforms the baseline
  algorithms.
\item We show the importance of different amino acids within the spike
  sequence by computing information gain corresponding to the class
  label.
\item We show that given the complexity of the data, our model is
  still able to outperform the baselines while using only 10\% of the
  training data.
\item We show that preserving the order of amino acids using $k$-mers
  achieves better predictive performance than the traditional one-hot
  encoding based embedding approach.
\item Our approach allows us to predict the geographical region of the
  COVID-19 infected patient while accounting for important local and
  global variability in the spike sequences.
\end{enumerate}

The rest of the paper is organized as follows:
Section~\ref{sec_related_work} contains the related work.  The
proposed approach is given in Section~\ref{sec_proposed_approach}.
Dataset detail and experimental setup are in
Section~\ref{sec_experimental_setup}.  The results of our method and
comparison with baselines is shown in
Section~\ref{sec_results_and_discussion}.  Finally, we conclude our
paper in Section~\ref{sec_conclusion}.

\section{Related Work}
%----------------------------------------------------------------------
\label{sec_related_work}

Sequence classification is a widely studied problem in domains like
sequence homology (shared ancestry) detection between a pair of
proteins and Phylogeny based inference~\cite{Dhar2020TNet} of disease
transmission~\cite{Krishnan2021PredictingVaccineHesitancy}.  Knowledge
of variants and mutations can also help in identifying the
transmission patterns of different variants, which will help to devise
appropriate public health interventions so that the rapid spread of
viruses can be
prevented~\cite{Ahmad2016AusDM,ahmad2017spectral,Tariq2017Scalable,AHMAD2020Combinatorial}.
This will also help in vaccine design and efficacy.  Previous studies
on working with a fixed length numerical representation of the data
successfully perform different data analytics tasks. It has
applications in different domains such as
graphs~\cite{hassan2020estimating,Hassan2021Computing}, nodes in
graphs~\cite{ali2021predicting,grover2016node2vec}, and electricity
consumption~\cite{ali2019short,Ali2020ShortTerm}.  This vector-based
representation also achieves significant success in sequence analysis,
such as
texts~\cite{shakeel2020multiDataAugmentation,Shakeel2020LanguageIndependent,shakeel2019multiBilingualSmsClassification},
electroencephalography and electromyography
sequences~\cite{atzori2014electromyography,ullah2020effect},
Networks~\cite{Ali2019Detecting1}, and biological
sequences~\cite{ali2021simpler}. However, most of the existing
sequence classification methods require the input sequences to be
aligned. Although sequence alignment helps to analyze the data better,
it is a very costly process.

In the evolution of the SARS-CoV-2 genome, it is well-known that a
disproportionate amount (in terms of its length) of the variation
takes place in the spike region.  Kuzmin \ea{}
in~\cite{kuzmin2020machine} show that viral-host classification can be
done efficiently using spike sequences only and applying different
machine learning (ML) models. They use one-hot encoding (OHE) to
obtain a numerical representation for the spike sequences and then
apply traditional ML classifiers after reducing the dimensions of the
data using the principal component analysis (PCA)
method~\cite{wold1987principal1}. Although OHE is proven to be
efficient in terms of predictive performance, it does not preserve the
order of amino acids in the spike protein if we want to take the
pair-wise Euclidean distance~\cite{ali2021k}. Another problem with the
one-hot encoding based approach is that it deals with aligned
sequential data only.

Many previous studies propose the use of $k$-mers (substrings of
length $k$), which is an alignment-free approach, instead of the
traditional OHE based embedding to obtain the numerical vector
representation for the genomic
data~\cite{ali2021effective,ali2021k,ali2021spike2vec}.  After
computing substrings of length $k$, a fixed-length feature vector is
generated, containing the count of each unique $k$-mer in a given
sequence. This $k$-mers based method has been used for phylogenetic
applications~\cite{Blaisdell1986AMeasureOfSimilarity} and has shown
success in constructing accurate phylogenetic trees from DNA
sequences.  Authors in~\cite{ali2021k} argue that better sequence
classification results can be achieved using $k$-mers instead of OHE
because $k$-mers tends to preserve the order of amino acids within a
(\eg, spike) sequence.

After obtaining the numerical representation, a popular approach is to
compute the kernel matrix and provide that matrix as input to
traditional machine learning classifiers like support vector machines
(SVM)~\cite{leslie2002mismatch,farhan2017efficient,Kuksa_SequenceKernel}.
Farhan \ea{} in~\cite{farhan2017efficient} propose an approximate
kernel (Gram matrix) computation algorithm, which uses the $k$-mers
based feature vector representation as an input to the kernel
computation algorithm.

\section{Proposed Approach}
%----------------------------------------------------------------------
\label{sec_proposed_approach}

In this section, we present our proposed model for classifying
population regions based on spike sequences only. We start by
explaining the basic MAJORITY based model for the classification. We
then show the one-hot encoding (OHE) based feature vector generation
approach. After that, we show how we generate $k$-mers based frequency
vectors. Then, we introduce our models, which we are using for the
purpose of classification.  Finally, we give brief details on the
experimental setup, before reporting the results of these experiments
in the following section.

\subsection{MAJORITY}

We start with a simple baseline model called MAJORITY. In this
approach, we simply take the class with majority representation in the
training data and declare it as the class label for all data points in
the test set. We then measure the performance of this baseline model
using different evaluation metrics.

\subsection{One-Hot Encoding~\cite{kuzmin2020machine}}

In order to obtain a numerical representation for the sequence-based
data, one of the popular methods is using one-hot encoding
(OHE)~\cite{kuzmin2020machine,ali2021effective,ali2021k,ali2021spike2vec}. Note
that the length of each spike sequence in our dataset is $1274$, which
contains characters (amino acids) from a set of $21$ unique alphabets
``\textit{ACDEFGHIKLMNPQRSTVWXY}". For OHE, since we need to have a
$21$ dimensional sub-vector for each amino acid, the length of the OHE
based feature vector for each spike sequence will be $21 \times 1273 =
26,733$ (we take the length of spike protein as $1273$ instead of
$1274$ because we have the stopping character `*' at the $1274^{th}$
position). After obtaining the OHE for the whole data, since the
dimensionality of the data will be high, authors
in~\cite{kuzmin2020machine} use the typical principal component
analysis (PCA) approach for dimensionality reduction. Since the size
of the data is much larger in our case, we simply cannot use PCA
because of high computational cost~\cite{ali2021spike2vec}. For this
purpose, we use instead an unsupervised approach for low dimensional
feature vector representation, called random Fourier features
(RFF)~\cite{rahimi2007random}.

\subsection{Random Fourier Features (RFF) Based Embedding~\cite{rahimi2007random}}

To compute the pair-wise similarity between two feature vectors, a
popular method is to compute the kernel (similarity) matrix (Gram
matrix) and give it as input to popular classifiers such as support
vector machine (SVM)~\cite{farhan2017efficient}. However, exact kernel
methods are expensive in terms of computation (scale poorly on
training data~\cite{rahimi2007random}), and they require huge space to
store an $n \times n$ matrix (where $n$ is the total number of data
points). To overcome this problem, we can use the so-called kernel
trick.

\begin{definition}[Kernel Trick]
  The kernel trick is a fast way to compute the similarity between
  feature vectors using the inner product. The kernel trick's main
  goal is to avoid the explicit need to map the input data to a
  high-dimensional feature space.
\end{definition}

The kernel trick relies on the assumption that any positive definite
function $f(a,b)$, where $a,b \in \mathcal{R}^d$, defines an inner
product and a lifting $\phi$ so that we can quickly compute the inner
product between the lifted data points~\cite{rahimi2007random}. It can
be described in a formal way as $\langle \phi (a), \phi (b) \rangle =
f(a,b)$.

Although the kernel trick is effective in terms of computational
complexity, it is still not scalable to millions of data points.  To
overcome these computational and storage problems, we use
RFF~\cite{rahimi2007random}, an unsupervised approach that maps the
input data to a randomized low dimensional feature space (Euclidean
inner product space). It can be described in a formal way as $z:
\mathcal{R}^d \rightarrow \mathcal{R}^D$.  In RFF, we approximate the
inner product between a pair of transformed points, which is almost
equal to the actual inner product between the original data
points. More formally: $f(a,b) = \langle \phi (a), \phi (b) \rangle
\approx z(a)' z(b)$.  Here, $z$ is a (transformed) low dimensional
(approximate) representation of the original feature vector (unlike
the lifting $\phi$). Since $z$ is the approximate low dimensional
representation of the original feature vector, we can use $z$ as an
input for different machine learning (ML) tasks such as
classification.

\subsection{Spike2Vec~\cite{ali2021spike2vec}}

Spike2Vec is a recently proposed method that uses $k$-mers and RFF to
design a low dimensional feature vector representation of the data and
then perform typical ML tasks such as classification and
clustering~\cite{ali2021spike2vec}. The first step of Spike2Vec is to
generate $k$-mers for the spike sequences.

\subsubsection{$k$-mers Computation}

\begin{figure}[!ht]
  \centering
  \includegraphics[scale = 0.4] {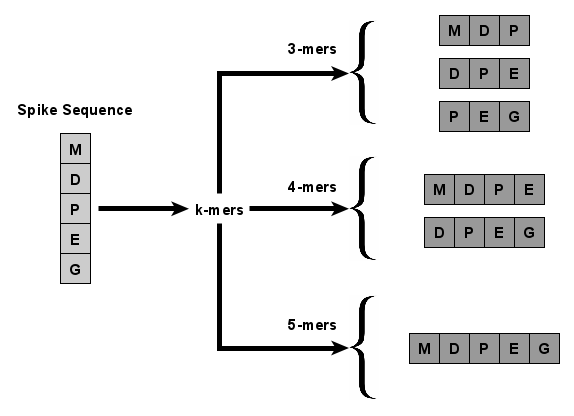}
  \caption{Example of different length $k$-mers in spike sequence
    ``MDPEG".}
  \label{fig_k_mer_demo}
\end{figure}

The main idea behind $k$-mers is to preserve the order of amino acids
within spike sequences. The $k$-mers is basically a set of substrings
(called mers) of length $k$. For each spike sequence, the total number
of $k$-mers is $N - k + 1$, where $N$ is the length of the spike
sequence ($1274$), and $k$ is a user-defined parameter for the size of
each mer.  An example of $k$-mers (where k = $3,4,$ and $5$) is given
in Figure \ref{fig_k_mer_demo}. In this paper, we are using $k = 3$
(selected empirically).

\subsection{Machine Learning Models}

To both feature engineering based embeddings, namely OHE and
Spike2Vec, we apply three ML classifiers downstream, namely Naive
Bayes (NB), Logistic Regression (LR), and Ridge Classifier (RC). For
all these classifiers, default parameters are used for training. To
measure the performance, we use average accuracy, precision, recall,
weighted and macro $F_1$, receiver operating characteristic area under
the curve (ROC-AUC). We also show the training runtime (in seconds)
for all methods.

\subsection{Neural Network Model}

Although the Spike2vec embedding allows the downstream ML models to
scale to datasets with millions of sequences, and is proven to
outperform the typical OHE, it is not always effective in terms of
overall predictive performance in certain scenarios. To further
increase the predictive performance, we move to a neural network (NN)
architecture, which takes OHE or $k$-mers based vectors as input.
Note that no dimensionality reduction step (\eg, PCA, RFF) is applied
beforehand --- the NN model takes the OHE or $k$-mers based vectors
directly.  Our NN architecture comprises a sequential constructor. We
create a fully connected network with one hidden layer that contains
$9261$ neurons (which is equal to the length of the feature vector
\ie, one neuron for every feature at the beginning). The activation
function that we are using is ``rectifier''. In the output layer, we
use the ``softmax'' activation function. At the end, we use the
efficient Adam gradient descent optimization
algorithm~\cite{zhang2018improved} with the ``sparse categorical
crossentropy'' loss function (used for multi-class classification
problems), which computes the crossentropy loss between the labels and
predictions. The batch size that we are taking is $100$, and we take
$10$ as the number of epochs for training the model. Note that we use
OHE and $k$-mers based frequency vectors (separately) as input to the
NN.

\begin{remark}
  Note that we are using ``sparse categorical crossentropy'' rather
  than simple ``categorical crossentropy'' because we are using
  integer labels rather than the one-hot representation of labels.
\end{remark}

%\textcolor{red}{\subsection{Gated Recurrent Units (GRU)} Another
%model that we tried is Gated Recurrent Units (GRU), a gating
%mechanism in recurrent neural networks. GRU is like long short-term
%memory (LSTM) with a forget gate,[2] but has fewer parameters than
%LSTM. It can only control information inside the unit and has no
%additional memory cell to keep information. Here, a one-layer
%sequential model with a GRU layer stacked on the top of a Dense layer
%is used as a predictive model. The activation function for the GRU
%layer is LeakyReLU, and for the Dense layer, we used the sigmoid
%function as an activation function. Also, the number of cells in the
%Dense layer equals the number of features in the data. Before
%training the model, we configure the learning process by compilation
%method that contains “Adam” as an optimizer, “sparse categorical
%crossentropy” loss function, and “accuracy” as a metric. A batch size
%of 100 samples from the training dataset estimates the error gradient
%before the model weights are updated, and we considered five epochs
%to train the model.}

\section{Experimental Evaluation}
%----------------------------------------------------------------------
\label{sec_experimental_setup}

In this section, we provide some statistics and visualization on the
data that we use, and then the precise details of the experimental
setup used to produce the results.

\subsection{Dataset Statistics}

We use a set of 2,384,646 spike amino acid sequences obtained from the
GISAID~\cite{gisaid_website_url}, along with metatdata on geographical
location (continent, country, and in the case of the USA, states).
This data, organized by country, is given in
Table~\ref{tbl_dataset_information}.

\begin{table}[ht!]
  \centering
  \resizebox{\textwidth}{!}{
    \begin{tabular}{cccccc}
      \midrule
      Region   & Country   & No. of sequences   & Region  & Country  & No. of sequences \\
      \toprule	\toprule	
      \multirow{19}{*}{Europe}  & England & 568202  & \multirow{2}{*}{North America}  & USA  & 663527\\
      & Germany  & 146730   & & Canada  & 91193 \\
      & Denmark  & 138574   & & Mexico  & 20040  \\
      \cmidrule{4-6}
      & Sweden  & 78810  & Total  & 3   & 774760 \\ 
      \cmidrule{4-6}\morecmidrules\cmidrule{4-6}
      & Scotland  & 69387   & \multirow{1}{*}{South America}  & Brazil   & 26729 \\
      \cmidrule{4-6}
      & France   & 56247   & Total  & 1   & 26729 \\
      \cmidrule{4-6}\morecmidrules\cmidrule{4-6}
      & Netherlands   & 49938 & \multirow{3}{*}{Asia}  & Japan  & 75423 \\
      & Spain   & 48830  & & India  & 37943 \\ 
      & Switzerland  & 48516  &   & Israel   & 14361 \\ 
      \cmidrule{4-6}
      & Wales   & 46851   & Total   & 3  & 127727\\
      \cmidrule{4-6}\morecmidrules\cmidrule{4-6}
      & Italy   & 44728  & \multirow{1}{*}{Australia}  & Australia  & 20985 \\
      \cmidrule{4-6}
      & Belgium  & 28758  & Total  & 1  & 20985 \\
      \cmidrule{4-6}
      & Ireland  & 23441 \\
      & Poland   & 16061 \\
      & Norway   & 14684 \\
      & Lithuania  & 13586 \\
      & Luxembourg  & 12713 \\
      & Finland  & 11254 \\
      & Slovenia & 17135 \\
      \cmidrule{1-3}
         Total  & 19  & 1434445 \\         
         \bottomrule
    \end{tabular}
  }
  \caption{The set of 2,384,646 SARS-CoV-2 spike sequences used in
    this study, labeled by country of origin.}
  \label{tbl_dataset_information}
\end{table}

\subsection{Data Visualization}

\begin{figure}[ht!]
  \centering
  \includegraphics[scale=0.5]{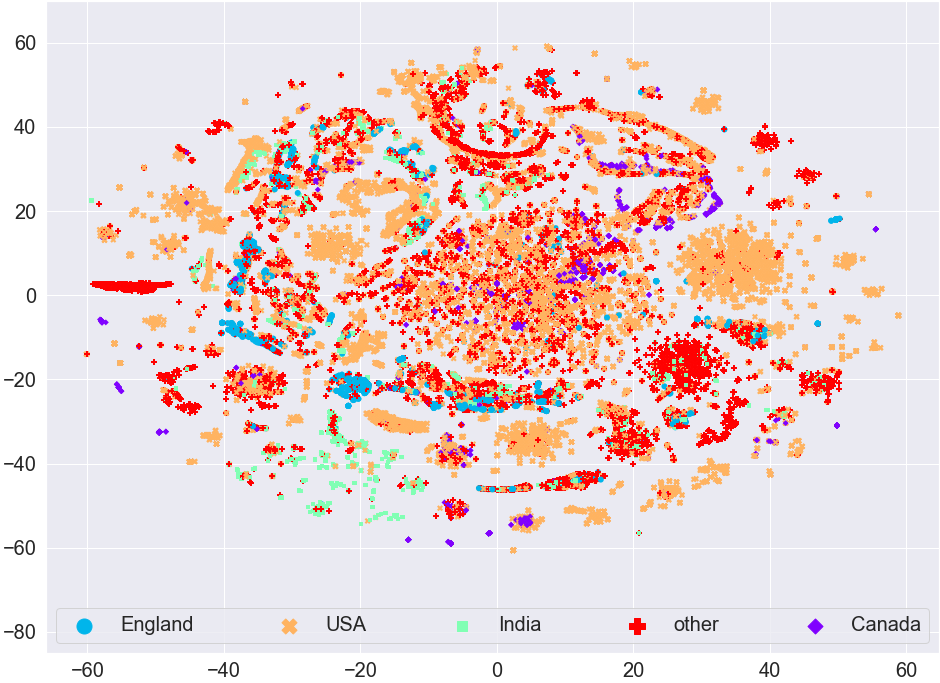}
  \caption{A t-SNE plot from the frequency ($k$-mers based feature)
    vectors along with the country information for $\approx$80K
    randomly sampled sequences from the set of 2,384,646 sequences
    (Table~\ref{tbl_dataset_information}) used in this study.}
  \label{fig_country_tsne_dataset_plot}
\end{figure}

To evaluate any natural clustering in our data (if any exist), we use
the t-distributed stochastic neighbor embedding
(t-SNE)~\cite{van2008visualizing}. The t-SNE approach maps the data
into a 2-dimensional (2D) real vector, which can then be visualized
using a scatter plot. Since applying t-SNE on the whole data is very
costly and time consuming, we randomly sampled a subset ($\approx$80K
sequences) from the data (of Table~\ref{tbl_dataset_information}) and
generated a 2D real vector using the t-SNE approach (see
Figure~\ref{fig_country_tsne_dataset_plot}).

\begin{remark}
  The reason for (randomly) selecting $\approx$80K sequences is
  because the t-SNE method is computationally very expensive (runtime
  is $O(N^2)$, where $N$ is the number of
  data-points~\cite{pezzotti2016approximated}) and is infeasible in
  terms of runtime on 2.3 million sequences.
\end{remark}

\begin{figure}[ht!]
  \centering
  \includegraphics[scale = 0.4] {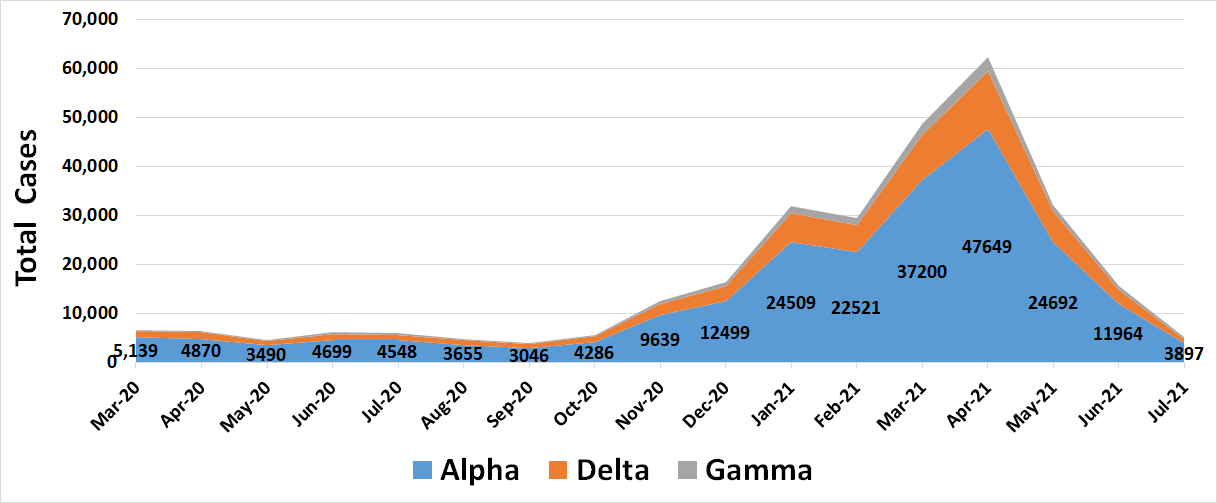}
  \caption{The rate of spread of the $3$ most common SARS-CoV-2
    variants (in the USA) from March $2020$ till July $2021$.}
  \label{fig_variant_distribution}
\end{figure}

The rate of spread of the $3$ most common variants of SARS-CoV-2 (in
the USA) from March $2020$ to July $2021$ from our data are given in
Figure~\ref{fig_variant_distribution}. We can see that the Alpha
variant
%(also known as UK variant~\cite{ali2021spike2vec})
was clearly the variant of concern when it reached its peak in April
$2021$. We can see a drop from this peak for all variants after April
$2021$. This is likely because a significant proportion of the
population were vaccinated by this point, hence the total number of
cases started decreasing~\cite{ali2021spike2vec}.

\subsection{Experimental Setup}

All experiments are conducted using an Intel(R) Xeon(R) CPU E7-4850 v4
@ $2.10$GHz having Ubuntu $64$ bit OS ($16.04.7$ LTS Xenial Xerus)
with 3023 GB memory.  The implementation of our algorithms is done in
Python and the code is available online for
reproducibility~\footnote{https://github.com/sarwanpasha/COVID-19-Country-Classification}.
% \footnote{Available at \url{https://github.com/sarwanpasha/COVID-19-Country-Classification}}.
We obtain a set of 2,384,646 spike amino acid sequences from the
GISAID~\cite{gisaid_website_url}.  The GISAID provides many different
metadata for these sequences, such as collection date, geographical
location, and sometimes variant information.  
% This data, preprocessed
% to include geographical location, is available
% online~\footnote{Available in published version},
% \footnote{Available at \url{https://drive.google.com/drive/folders/1-YmIM8ipFpj-glr9hSF3t6VuofrpgWUa?usp=sharing}},
which can be used after agreeing to the terms and conditions of
GISAID\footnote{Available at \url{https://www.gisaid.org/}}.
% Authors in~\cite{ali2021k} performs variant classification on COVID-19 data using kernel approach~\cite{farhan2017efficient}. However, they use a minimal sized dataset (the biggest dataset has $7000$ spike sequences). One major problem with the kernel method used in~\cite{ali2021k} is that it is not scalable on bigger datasets~\cite{farhan2017efficient}. Although
For the classification algorithms, we use $10\%$ of the data for
training and $90\%$ for testing. The purpose of using a smaller
training dataset is to show how much performance gain we can achieve
while using minimal training data.
% \subsection{Dataset Statistics}

% \begin{remark}
%   Our data split and pre-processing follow those of~\cite{ali2021k,ali2021spike2vec,ali2021effective}.
% \end{remark}

\section{Results and Discussion}
%----------------------------------------------------------------------
\label{sec_results_and_discussion}

In this section, we present results for three different granularities
of class labels, namely continents, countries, and finally states in a
case study of the United States of America (USA).

\subsection{Continent Classification}

In this section, we show classification results for $5$ different
continents, namely Europe, North America, South America, Asia, and
Australia (see Table~\ref{tbl_dataset_information}). The
classification results (average $\pm$ standard deviation of 5 runs)
are given in Table~\ref{tbl_continent_classification_results}. In
terms of predictive performance, we can observe that the NN model with
the $k$-mers based embedding performs best compared to the
baselines. While comparing the two embedding methods (\ie, OHE and
$k$-mers), we can see that $k$-mers is better than OHE for the NN
model.
%\textcolor{red}{GRU} and other ML algorithms.
Since $k$-mers can preserve the order of amino acids better as
compared to the OHE, it is able to give richer information in the
feature vector. In terms of runtime, RC with the Spike2Vec embedding
is performing best. The NN model will take longer to train the models
compared to simple ML classifiers because of the tuning of different
parameters.

\begin{table}[!ht]
  \centering
  \resizebox{1\textwidth}{!}{
    \begin{tabular}{p{1.9cm}ccp{2.2cm}p{2.2cm}p{2.2cm}p{2.2cm}p{2.2cm}p{2.2cm} | c}
      \midrule
      \multirow{3}{*}{Approach} & \multirow{3}{1.1cm}{Embed. Method} & \multirow{3}{*}{Algo.} & \multirow{3}{*}{Accuracy}  & \multirow{3}{*}{Precision}  & \multirow{3}{*}{Recall}  & \multirow{3}{0.9cm}{$F_1$ weigh.}  & \multirow{3}{0.9cm}{$F_1$ Macro}  & \multirow{3}{0.9cm}{ROC-AUC}  & \multirow{3}{2.2cm}{Training runtime (sec.)} \\
      &&&&&&&&& \\
      &&&&&&&&& \\
      \toprule	\toprule	
      \multirow{1}{*}{MAJORITY} & \_  & \_  & 0.60 $\pm$ 0.000  & 0.36 $\pm$ 0.000  & 0.60 $\pm$ 0.000  & 0.45 $\pm$ 0.000  & 0.15 $\pm$ 0.000  & 0.50 $\pm$ 0.000 &  \_ \\
      \midrule
      \multirow{6}{1.9cm}{Feature Engineering} & \multirow{3}{*}{OHE}  
      & NB & 0.49 $\pm$ 0.005  & 0.63  $\pm$ 0.006 & 0.49  $\pm$ 0.005 & 0.50  $\pm$ 0.007 & 0.38  $\pm$ 0.006 & 0.63  $\pm$ 0.005 &  1457.2 $\pm$ 0.023 \\
      & & LR & 0.67  $\pm$ 0.007 & 0.66  $\pm$ 0.008 & 0.67  $\pm$ 0.007 & 0.64  $\pm$ 0.007 & 0.33  $\pm$ 0.008 & 0.58  $\pm$ 0.005 &  1622.4 $\pm$ 0.031 \\
      & & RC & 0.67  $\pm$ 0.004 & 0.66  $\pm$ 0.005 & 0.67  $\pm$ 0.004 & 0.64  $\pm$ 0.006 & 0.28  $\pm$ 0.004 & 0.57  $\pm$ 0.005 &  1329.1 $\pm$ 0.029 \\
      \cmidrule{2-10}
      & \multirow{3}{*}{Spike2Vec}  
      & NB  & 0.48  $\pm$ 0.007 & 0.63  $\pm$ 0.006 & 0.48  $\pm$ 0.008 & 0.49  $\pm$ 0.007 & 0.36  $\pm$ 0.007 & 0.63  $\pm$ 0.006 &  970.6 $\pm$ 0.065 \\
      & & LR &  0.67 $\pm$ 0.005 & 0.67  $\pm$ 0.007 & 0.67  $\pm$ 0.006 & 0.64  $\pm$ 0.007 & 0.34  $\pm$ 0.006 & 0.58  $\pm$ 0.005 &  1141.9 $\pm$ 0.072 \\
      & & RC &  0.67  $\pm$ 0.003 & 0.66  $\pm$ 0.004 & 0.67  $\pm$ 0.003 & 0.64  $\pm$ 0.003 & 0.29  $\pm$ 0.006 & 0.57  $\pm$ 0.007 &  \textbf{832.3} $\pm$ 0.057 \\
      \midrule
      \multirow{4}{*}{NN} & \multirow{2}{*}{OHE}  
      & \multirow{2}{1.3cm}{Neural Network}  &  \multirow{2}{*}{0.75 $\pm$ 0.007}  & \multirow{2}{*}{0.76 $\pm$ 0.008}  & \multirow{2}{*}{0.75 $\pm$ 0.008}  & \multirow{2}{*}{0.72 $\pm$ 0.009}  & \multirow{2}{*}{0.47 $\pm$ 0.007}  & \multirow{2}{*}{0.65 $\pm$ 0.008}  & \multirow{2}{*}{30932.0 $\pm$ 0.105} \\
      &&&&&&&&& \\
      %& $\pm$ & \textcolor{red}{GRU} $\pm$ & \textcolor{red}{0.60}
      %$\pm$ & \textcolor{red}{0.36} $\pm$ & \textcolor{red}{0.60}
      %$\pm$ & \textcolor{red}{0.45} $\pm$ & \textcolor{red}{0.15}
      %$\pm$ & \textcolor{red}{0.5} $\pm$ & \textcolor{red}{55072.81}
      %\\
      \cmidrule{2-10}
      & \multirow{2}{*}{$k$-mers}  
      & \multirow{2}{1.3cm}{Neural Network}  &  \multirow{2}{*}{\textbf{0.77} $\pm$ 0.009}  & \multirow{2}{*}{\textbf{0.78} $\pm$ 0.008} & \multirow{2}{*}{\textbf{0.77} $\pm$ 0.009}  & \multirow{2}{*}{\textbf{0.74} $\pm$ 0.007}  & \multirow{2}{*}{\textbf{0.49} $\pm$ 0.008}  & \multirow{2}{*}{\textbf{0.65} $\pm$ 0.009}  &  \multirow{2}{*}{18631.7 $\pm$ 0.235} \\
      &&&&&&&&& \\
      \bottomrule
    \end{tabular}
  }
  \caption{Continent classification results (average $\pm$ standard
    deviation of 5 runs) for $5$ continents comprising 2,384,646 spike
    sequences (10\% training set and 90\% testing set). Best average
    values are shown in bold.}
  \label{tbl_continent_classification_results}
\end{table}

\subsection{Country Classification}

After classifying the continents, we take countries as the class label
and train all ML and NN models again with the same parameter
settings. The classification results (average $\pm$ standard deviation
of 5 runs) for countries is given in
Table~\ref{tbl_classification_results}. In terms of predictive
performance, we can observe that the NN model is performing better
than all baselines. In terms of runtime, RC with the OHE is the best
classifier. An important observation here is the drop in overall
performance of all classification models as compared to the continent
classification. The reason for this behavior is likely due to any
natural clustering or other information in the spike sequences
corresponding to the location of patients breaking down at this level
of granularity.
%(see Figure~\ref{fig_country_tsne_dataset_plot}).
This lack of knowledge in the data makes country classification a
difficult task. However, we can see that the NN model can still
classify the countries better than the baselines.

% 598.481 (kernel approx runtime!!!!)
\begin{table}[!ht]
  \centering
  \resizebox{1\textwidth}{!}{
    \begin{tabular}{p{1.9cm}ccp{2.2cm}p{2.2cm}p{2.2cm}p{2.2cm}p{2.2cm}p{2.2cm} | c}
      \midrule
      \multirow{3}{*}{Approach} & \multirow{3}{1.1cm}{Embed. Method} & \multirow{3}{*}{Algo.} & \multirow{3}{*}{Accuracy}  & \multirow{3}{*}{Precision}  & \multirow{3}{*}{Recall}  & \multirow{3}{0.9cm}{$F_1$ weigh.}  & \multirow{3}{0.9cm}{$F_1$ Macro}  & \multirow{3}{0.9cm}{ROC-AUC}  & \multirow{3}{2.2cm}{Training runtime (sec.)} \\
      &&&&&&&&& \\
      &&&&&&&&& \\
      \toprule	\toprule
      \multirow{1}{*}{MAJORITY} & \_  & \_  & 0.27 $\pm$ 0.000 & 0.07 $\pm$ 0.000 &  0.27 $\pm$ 0.000 & 0.12 $\pm$ 0.000 & 0.01 $\pm$ 0.000 & 0.50 $\pm$ 0.000 &  \_ \\
      \midrule
      \multirow{6}{1.9cm}{Feature Engineering} & \multirow{3}{*}{OHE}  
      & NB  & 0.11 $\pm$ 0.007 & 0.44 $\pm$ 0.008 & 0.11 $\pm$ 0.007 & 0.11 $\pm$ 0.007 & 0.10 $\pm$ 0.009 &  0.55 $\pm$ 0.008 &  1308.4 $\pm$ 0.098 \\
      & & LR & 0.40 $\pm$ 0.009 & 0.46 $\pm$ 0.009 & 0.40 $\pm$ 0.008 & 0.33 $\pm$ 0.007 & 0.15 $\pm$ 0.008 & 0.55  $\pm$ 0.009 &  2361.8 $\pm$ 0.074 \\
      & & RC & 0.40 $\pm$ 0.006 & 0.38 $\pm$ 0.007 & 0.40 $\pm$ 0.006 & 0.31 $\pm$ 0.008 & 0.11 $\pm$ 0.007 & 0.54  $\pm$ 0.006 &  \textbf{746.4} $\pm$ 0.085 \\
      \cmidrule{2-10}
      & \multirow{3}{*}{Spike2Vec}  
      & NB & 0.13 $\pm$ 0.004 & 0.41 $\pm$ 0.005 & 0.13 $\pm$ 0.004 & 0.151 $\pm$ 0.006 & 0.109 $\pm$ 0.005 & 0.555 $\pm$ 0.007 &  1315.3 $\pm$ 0.085 \\
      & & LR & 0.40 $\pm$ 0.006 & 0.45 $\pm$ 0.006 & 0.40 $\pm$ 0.007 & 0.33 $\pm$ 0.008 & 0.16 $\pm$ 0.007 & 0.55  $\pm$ 0.006 &  2736.8 $\pm$ 0.058 \\
      & & RC & 0.39 $\pm$ 0.006 & 0.37 $\pm$ 0.007 & 0.39 $\pm$ 0.006 & 0.31 $\pm$ 0.008 & 0.11 $\pm$  0.006 & 0.54  $\pm$ 0.007 &  779.4 $\pm$ 0.074 \\
      \midrule
      \multirow{4}{*}{NN}  & \multirow{2}{*}{OHE}  
      & \multirow{2}{1.3cm}{Neural Network}  & \multirow{2}{*}{0.49 $\pm$ 0.009} & \multirow{2}{*}{0.53 $\pm$ 0.008} & \multirow{2}{*}{0.49 $\pm$ 0.009} & \multirow{2}{*}{0.43 $\pm$ 0.009} & \multirow{2}{*}{0.24 $\pm$ 0.007} & \multirow{2}{*}{0.6 $\pm$ 0.006} & \multirow{2}{*}{28914.8 $\pm$ 0.453} \\
      &&&&&&&&& \\
      %&  $\pm$ & \textcolor{red}{GRU} $\pm$ &  \textcolor{red}{0.0} $\pm$ & \textcolor{red}{0.0} $\pm$ & \textcolor{red}{0.0} $\pm$ & \textcolor{red}{0.0} $\pm$ & \textcolor{red}{0.0} $\pm$ & \textcolor{red}{0.0} $\pm$ & \textcolor{red}{0.0} \\
      \cmidrule{2-10}
      & \multirow{2}{*}{$k$-mers}  &
      \multirow{2}{1.3cm}{Neural Network}  & \multirow{2}{*}{\textbf{0.51} $\pm$  0.005} & \multirow{2}{*}{\textbf{0.57} $\pm$ 0.004} & \multirow{2}{*}{\textbf{0.51} $\pm$ 0.005} & \multirow{2}{*}{\textbf{0.45} $\pm$  0.006} & \multirow{2}{*}{\textbf{0.28} $\pm$  0.006} & \multirow{2}{*}{\textbf{0.60} $\pm$ 0.007} & \multirow{2}{*}{10383.6 $\pm$ 0.745} \\
      &&&&&&&&& \\
      %&  $\pm$ & \textcolor{red}{GRU} $\pm$ &  \textcolor{red}{0.0} $\pm$ & \textcolor{red}{0.0} $\pm$ & \textcolor{red}{0.0} $\pm$ & \textcolor{red}{0.0} $\pm$ & \textcolor{red}{0.0} $\pm$ & \textcolor{red}{0.0} $\pm$ & \textcolor{red}{0.0} \\
      \bottomrule
    \end{tabular}
  }
  \caption{Country classification results (average $\pm$ standard
    deviation of 5 runs) for $27$ countries comprising 2,384,646 spike
    sequences (10\% training set and 90\% testing set). Best average
    values are shown in bold.}
  \label{tbl_classification_results}
\end{table}

\subsection{A Case Study of the United States of America (USA)}

\begin{figure}[h!]
  \centering
  \includegraphics[scale = 0.6]{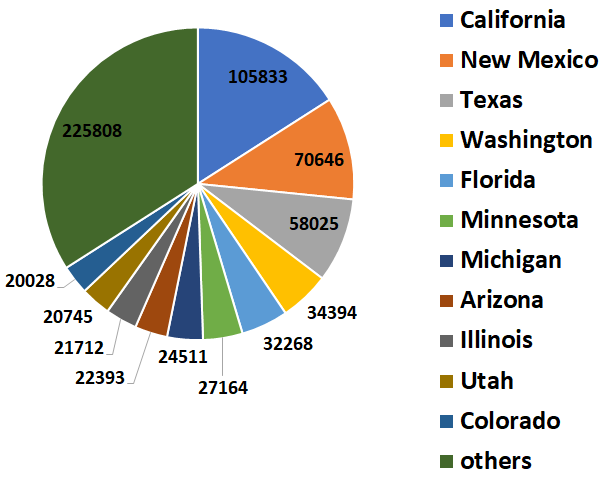}
  \caption{Distribution of the 663,527 sequences over the states of
    the USA, with the top 11 states specified, while the remaining
    fall into the ``others'' category.}
  \label{fig_usa_states}
\end{figure}

After classifying continents and countries, we investigate our model
with more highly granular class labels. For this purpose, we first
take the single country with the highest number of spike sequences in
the data. Since the USA contains most of the spike sequences in our
data (see Table~\ref{tbl_dataset_information}), we took it as a case
study to further explore different states within the USA. The pie
chart showing the distribution of the sequences over the states of the
USA is given in Figure~\ref{fig_usa_states}. The classification
results (average $\pm$ standard deviation of 5 runs) for different
states are given in Table~\ref{tbl_usa_states_results}. We can again
observe the drop in predictive performance for all
%ML and Deep learning
models. This again proves that as we increase the granularity of the
class labels, it becomes difficult for any model to classify with
higher accuracy. We can also observe that the NN model with the
$k$-mers based feature embedding is performing better than all the
baselines.

\begin{table}[!ht]
  \centering
  \resizebox{1\textwidth}{!}{
    \begin{tabular}{p{1.9cm}ccp{2.2cm}p{2.2cm}p{2.2cm}p{2.2cm}p{2.2cm}p{2.2cm} | c}
      \midrule
      \multirow{3}{*}{Approach} & \multirow{3}{1.1cm}{Embed. Method} & \multirow{3}{*}{Algo.} & \multirow{3}{*}{Accuracy}  & \multirow{3}{*}{Precision}  & \multirow{3}{*}{Recall}  & \multirow{3}{0.9cm}{$F_1$ weigh.}  & \multirow{3}{0.9cm}{$F_1$ Macro}  & \multirow{3}{0.9cm}{ROC-AUC}  & \multirow{3}{2.2cm}{Training runtime (sec.)} \\
      &&&&&&&&& \\
      &&&&&&&&& \\
      \toprule	\toprule
      \multirow{1}{*}{MAJORITY} & \_  & \_  & 0.33 $\pm$ 0.000 & 0.11 $\pm$ 0.000 & 0.33 $\pm$ 0.000 & 0.17 $\pm$ 0.000 & 0.04 $\pm$ 0.000 & 0.50 $\pm$ 0.000 &  \_ \\
      \midrule
      \multirow{6}{1.9cm}{Feature Engineering} & \multirow{3}{*}{OHE}  
      & NB & 0.18 $\pm$ 0.005 & 0.32 $\pm$ 0.006 & 0.18 $\pm$ 0.004 & 0.14 $\pm$ 0.005 & 0.13 $\pm$ 0.006 & 0.54 $\pm$ 0.007 & 860.2 $\pm$ 0.745 \\
      &  & LR & 0.37 $\pm$ 0.007 & 0.45 $\pm$ 0.008 & 0.37 $\pm$ 0.007 & 0.26 $\pm$ 0.007 & 0.13 $\pm$ 0.008 & 0.53 $\pm$ 0.006 & 1036.2 $\pm$ 0.458 \\
      &  & RC & 0.37 $\pm$ 0.008 & 0.41 $\pm$ 0.008 & 0.37 $\pm$ 0.007 & 0.25 $\pm$ 0.009 & 0.12 $\pm$ 0.005 & 0.52 $\pm$ 0.007 & 707.7 $\pm$ 0.865 \\
      \cmidrule{2-10}
      & \multirow{3}{*}{Spike2Vec}  
      & NB & 0.19 $\pm$ 0.005 & 0.37 $\pm$ 0.005 & 0.19 $\pm$ 0.004 & 0.14 $\pm$ 0.006 & 0.14 $\pm$ 0.006 & 0.55 $\pm$ 0.007 &  273.7 $\pm$ 0.124 \\
      & & LR & 0.38 $\pm$ 0.008 & 0.44 $\pm$ 0.007 & 0.38 $\pm$ 0.007 & 0.29 $\pm$ 0.006 & 0.16 $\pm$ 0.007 & 0.54 $\pm$ 0.008 & 374.2 $\pm$  0.865 \\
      &  & RC & 0.37 $\pm$ 0.006 & 0.42 $\pm$ 0.006 & 0.37 $\pm$ 0.004 & 0.27 $\pm$ 0.005 & 0.14 $\pm$ 0.008 & 0.53 $\pm$ 0.007 & \textbf{197.1} $\pm$ 0.657 \\
      \midrule
      \multirow{4}{*}{NN} & \multirow{2}{*}{OHE}  
      & \multirow{2}{1.3cm}{Neural Network}  & \multirow{2}{*}{0.38 $\pm$ 0.011} & \multirow{2}{*}{0.44 $\pm$ 0.012} & \multirow{2}{*}{0.38 $\pm$ 0.011} & \multirow{2}{*}{0.34 $\pm$ 0.013} & \multirow{2}{*}{0.22 $\pm$ 0.015} & \multirow{2}{*}{0.57 $\pm$ 0.010} &  \multirow{2}{*}{7881.3 $\pm$ 0.857} \\
      &&&&&&&&& \\
      %& $\pm$ & \textcolor{red}{GRU} $\pm$ &  \textcolor{red}{0.34} $\pm$ & \textcolor{red}{0.11} $\pm$ & \textcolor{red}{0.34} $\pm$ & \textcolor{red}{0.17} $\pm$ & \textcolor{red}{0.04} $\pm$ & \textcolor{red}{0.5} $\pm$ & \textcolor{red}{15829.47} \\
      \cmidrule{2-10}
      & \multirow{2}{*}{$k$-mers}  &
      \multirow{2}{1.3cm}{Neural Network}  & \multirow{2}{*}{\textbf{0.47} $\pm$ 0.017} & \multirow{2}{*}{\textbf{0.50} $\pm$ 0.016} & \multirow{2}{*}{\textbf{0.47} $\pm$ 0.014} & \multirow{2}{*}{\textbf{0.42} $\pm$ 0.017} & \multirow{2}{*}{\textbf{0.33} $\pm$ 0.015} & \multirow{2}{*}{\textbf{0.61} $\pm$ 0.017} & \multirow{2}{*}{4908.6 $\pm$ 0.975} \\
      &&&&&&&&& \\
      % \cmidrule{2-10}
      % $\pm$ &  $\pm$ & \textcolor{red}{GRU} $\pm$ &  \textcolor{red}{0.0} $\pm$ & \textcolor{red}{0.0} $\pm$ & \textcolor{red}{0.0} $\pm$ & \textcolor{red}{0.0} $\pm$ & \textcolor{red}{0.0} $\pm$ & \textcolor{red}{0.0} $\pm$ & \textcolor{red}{0.0} \\
      \bottomrule
      
    \end{tabular}
  }
  \caption{Classification results (average $\pm$ standard deviation of
    5 runs) for different states of the USA (10\% training set and
    90\% testing set). The best average values are shown in bold.}
  \label{tbl_usa_states_results}
\end{table}

\subsection{Importance of Attributes}
%----------------------------------------------------------------------
\label{sec_att_imp}

\begin{figure}[h!]
  \centering
  % \begin{subfigure}{.33\textwidth}
  \centering
  \includegraphics[scale = 0.8] {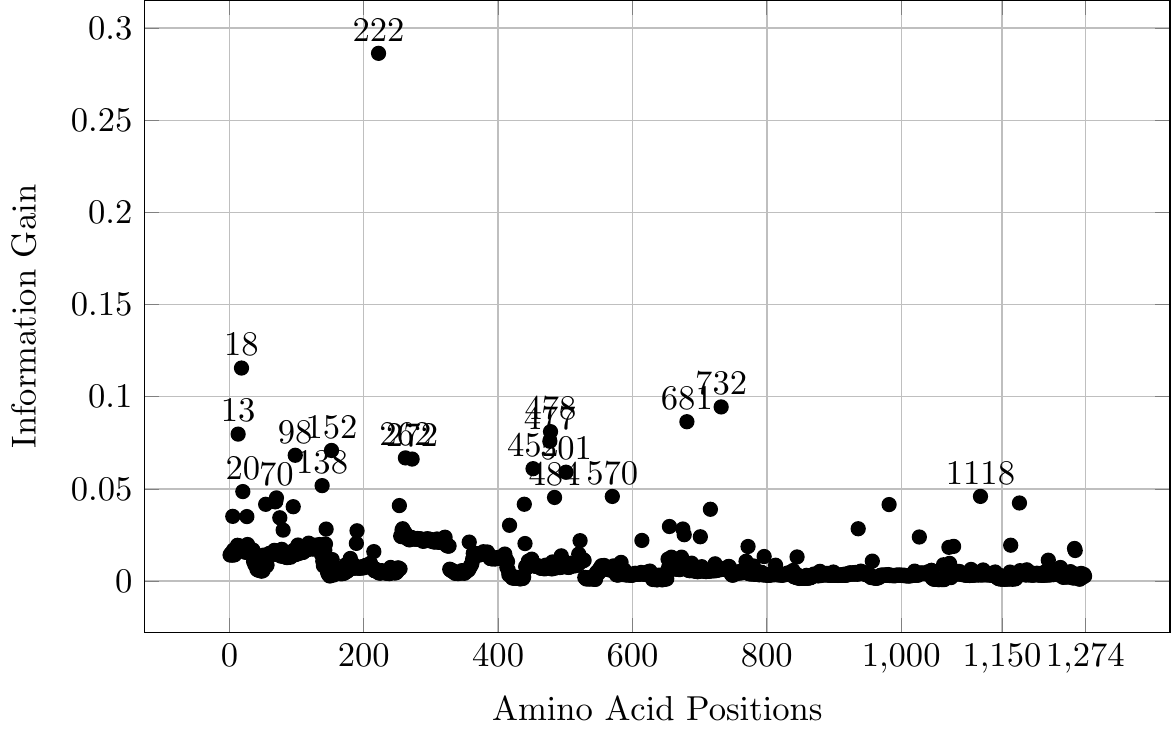}
  \caption{Information gain for each amino acid position corresponding
    to the class.}
  \label{fig_attribute_correlation}
\end{figure}

To evaluate the importance of the positions in the spike sequences, we
find the importance of each attribute with respect to class label
(using the Weka tool\footnote{Available at
  \url{https://www.cs.waikato.ac.nz/ml/weka/}}). For this purpose, a
randomly selected a subset of spike sequences ($\approx$80K) is taken
from the original dataset. We then compute the information gain (IG)
between each attribute ($P =$ amino acid position) and the true class
label ($C =$ country). More formally, IG can be computed as

\begin{equation}
  IG(C,P) = H(C) - H(C|P),
\end{equation}

\noindent where $H(C)$ and $H(C|P)$ are entropy and conditional
entropy, respectively. The entropy $H$ can be calculated using

\begin{equation}
  H = \sum_{ i \in C} -p_i \log p_i,
\end{equation}

\noindent where $p_i$ is the probability of the class $i$.  The IG
values for each attribute are given in
Figure~\ref{fig_attribute_correlation}.  The IG values for each
attribute are also available online\footnote{Available at
  \url{https://github.com/sarwanpasha/COVID-19-Country-Classification/blob/main/attributes_correlation.csv}}.

\section{Conclusion}
%----------------------------------------------------------------------
\label{sec_conclusion}

This paper employs several machine learning models using a $k$-mers
based representation as input and efficiently classifies SARS-CoV-2
spike sequences based on geographical location. We show that our
proposed approach outperforms the baselines in terms of predictive
performance. Using information gain, we also show the importance of
attributes (amino acids) in the spike sequences.  Such classification
and its analysis can help researchers to study more deeply the
connection between geographical location and SARS-CoV-2 variants.  In
the future, we will explore more sophisticated models like LSTM and
GRU, and also use other attributes like months information to increase
the predictive performance, and maybe give an idea of the dynamics
(spread) of the virus over time. Using other alignment-free methods
such as minimizers is another possible future direction.

\section*{Acknowledgements}

Paper was originally deposited to arXiv:
\url{https://arxiv.org/abs/2110.00809}

\section*{Author Contribution Statement}

\textbf{SA, MP:} Conceptualization. \textbf{SA:} Methodology.
\textbf{SA:} Software. \textbf{SA, ZT}: Validation. \textbf{SA, MP:}
Formal Analysis. \textbf{SA, BB, ZT:} Investigation. \textbf{All:}
Resources. \textbf{All:} Data Curation. \textbf{All:} Writing -
Original Draft. \textbf{All:} Writing - Review \&
Editing. \textbf{SA:} Visualization. \textbf{SA, MP:}
Supervision. \textbf{MP:} Project Administration. \textbf{MP:} Funding
Acquisition.

\section*{Conflict of Interest}

The authors declare no conflict of interest.

\section*{Funding Statement}

Research supported by an MBD Fellowship for SA, and a Georgia State
University Computer Science start-up grant for MP.

\bibliographystyle{plainnat}
\bibliography{references}

\end{document}